\documentclass[letterpaper, 10 pt, conference]{ieeeconf}  

\IEEEoverridecommandlockouts                              

\usepackage{graphics} 
\usepackage{amsmath} 
\usepackage{amssymb}  
\usepackage{todonotes}  
\usepackage{outlines}  
\usepackage{algorithm} 
\usepackage{algpseudocode} 
\usepackage{booktabs} 
\usepackage{bm}
\usepackage{subcaption} 
\usepackage{bbm} 
\usepackage{etoolbox}
\AtBeginEnvironment{tabular}{\footnotesize}

\usepackage{caption}
\usepackage{cite}

\title{\LARGE \bf
Planning-Query-Guided Model Generation for Model-Based Deformable Object Manipulation
}

\author{
Alex LaGrassa$^{1,*}$ and Zixuan Huang$^{2,*}$ and Dmitry Berenson$^{2}$ and Oliver Kroemer$^{3}$
\thanks{$^{1}$Department of Computer Science, Stevens Institute of Technology. Work performed at Robotics Institute, Carnegie Mellon University} 
\thanks{$^{2}$Department of Robotics, University of Michigan}%
\thanks{$^{3}$Robotics Institute, Carnegie Mellon University} 
\thanks{Correspondence to \texttt{alagrass@stevens.edu}}
\thanks{This work was supported in part by NSF Robust Intelligence under Grant 1956163, NSF/USDA NIFA AIIRA AI Research Institute Grant 2021-67021-35329, ONR Grant N00014-24-1-2036, and NSF Grants IIS-2113401 and IIS-2220876.}
}

\begin{document}

\maketitle
\thispagestyle{empty}
\pagestyle{empty}

\newcommand{\figwidth}{0.5\textwidth}

\begin{abstract}
Efficient planning in high-dimensional spaces, such as those involving deformable objects, requires computationally tractable yet sufficiently expressive dynamics models. 
This paper introduces a method that automatically generates \emph{task-specific, spatially adaptive dynamics models} by learning which regions of the object require high-resolution modeling to achieve good task performance \emph{for a given planning query}. 
Task performance depends on the complex interplay between the dynamics model, world dynamics, control, and task requirements. 
Our proposed diffusion-based \emph{model generator} predicts per-region model resolutions based on start and goal pointclouds that define the planning query. 
To efficiently collect the data for learning this mapping, a two-stage process optimizes resolution using predictive dynamics as a prior before directly optimizing using closed-loop performance. 
On a tree-manipulation task, our method doubles planning speed with only a small decrease in task performance over using a full-resolution model. This approach informs a path towards using previous planning and control data to generate computationally efficient yet sufficiently expressive dynamics models for new tasks. 
 
\end{abstract}
\def\vec{\mathaccent "017E\relax }

\section{Introduction}
\newcommand{\modelres}{\mathbf{\vec{\omega}}}
\newcommand{\bomega}{\mathbf{\omega}}

Deep learning on general-purpose data representations, such as images and point clouds, has enabled predictive world models of high-dimensional deformable objects with complex dynamics~\cite{zhang2024adaptigraph, chen2024differentiable}. Furthermore, scalable generative models trained on large, diverse datasets can broadly generalize across domains. However, computation on high-dimensional data comes at a cost. Many large models have slow inference times due to using billions of parameters. These computation times make them impractical for applications like model-based planning, which require repeated sequential evaluation for good performance. 

This paper focuses on graph neural networks (GNNs), commonly used for modeling deformable objects~\cite{huang2022mesh, allen2022learning, sanchez2020learning, zhang2024adaptigraph}, operate on graph-based representation whose size, defined by the number of vertices and edges, determines their \emph{resolution}. Higher-resolution graphs capture more detail at greater computational cost. State representations and corresponding models must strike a balance. They must be expressive enough to solve the task, but not so large that planning becomes computationally intractable~\cite{pfaff2020learning, fortunato2022multiscale}. GNNs are particularly powerful since they can generalize across highly different graph structures and node quantities~\cite{morrison2024gfn}.

\begin{figure}[ht!]
\includegraphics[width=0.44\textwidth]{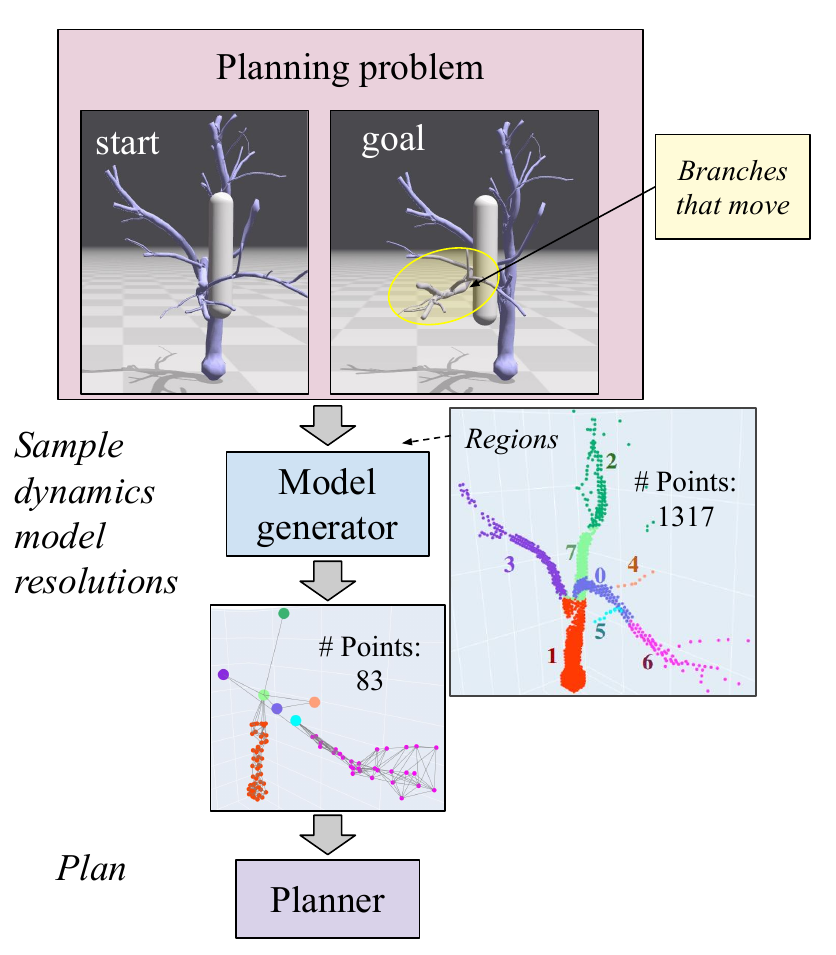}
\caption{Given a planning query and a segmented pointcloud, the model generator outputs a dynamics model that simplifies regions that are irrelevant to task performance. In this example, the circled branch moves between the start and goal. The model generator outputs a graph that represents the end of that branch and the base in high resolution, and simplifies all other regions. Although the tree has 1317 particles, only 227 particles are needed to represent the tree dynamics accurately. The simplified model is then used for faster planning. }
\label{fig:intuitive_example}
\end{figure}
This paper addresses the problem of simplifying predictive models without sacrificing end-to-end closed-loop task performance by learning which parts of the state needs high-resolution modeling and which can be safely simplified to a lower resolution. Task-relevant simplification depends on a complex interplay between the dynamics model, task objectives, underlying world dynamics, planner, and low-level controller. Consider a robot tasked with moving a particular tree branch to inspect or remove fruit from it (Fig.~\ref{fig:model_gen_visual_overview}). Intuitively, some regions, such as branches far from the apple, \emph{may} be modeled coarsely without harming performance. However, the required resolutions depend on factors such as if they need to be pushed or if they may be moved by other branches to solve the task. Resolution-adaptive representations from computer graphics are a helpful starting point for our work. However, region-specific resolution selection in planning depends on the \emph{ability to make good decisions} during planning, instead of the more directly optimizable geometric fidelity. In fact, prior work has shown that low-resolution models can be surprisingly effective for control in some cases~\cite{suh2021surprising, kamel2017linear}.

Our contributions are as follows: 
\begin{enumerate}
    \item A \textbf{diffusion-based model generator} that outputs region-specific resolutions that optimize computational efficiency for a given planning query with minimal impact to closed-loop task performance. The input is represented as a start and goal point cloud. The output is a vector that specifies how to encode the point cloud into a graph for planning. To train this model generator, we construct a dataset of planning queries paired with optimized resolution vectors
    \item A \textbf{two-stage resolution optimization procedure} for generating this training data. Since evaluating performance with closed-loop MPC is computationally expensive, the first stage produces a prior based on dynamics model accuracy, which is then refined using closed-loop task performance. 
\end{enumerate}


\section{Related Work}

\par \textbf{Task-Relevant Model Selection: }
The most relevant area of work to ours is model selection: selecting a dynamics model that a controller uses for optimization. Predictive accuracy is not necessary or even necessarily correlated with task performance. Only some parts of the state-space are relevant to the planning query. Some methods in search-based planning select the necessary DoF needed at various parts of the search to compute a plan to a specified goal~\cite{styler2017plan, youakim2018motion} or use a heuristic such as the presence of contacts to decide when to use a more accurate but slower physics model~\cite{saleem2020planning}. In high-dimensional problems such as deformable object manipulation, the relationship between predictive accuracy and planning performance is even more complex~\cite{suh2021surprising}. With access to a complete and accurate model, an assumption we do not make, some methods can directly optimize state-space and action-space reductions that preserve task performance while making computation tractable~\cite{wang2024learning, russell2024online, koessler2021efficient}.

Without accurate models, optimal model selection for task performance requires evaluating controller performance on the true dynamical system. Since evaluation on the true system can be expensive,~\cite{mcconachie2018estimating} frames model selection as a multi-arm bandit problem where the utility of a model is measured by how much a generated command reduces the current error to the goal. Although this work acknowledges the computational benefit of more simplified models, unlike in our method, computational cost is not reflected in the model selection process. However,~\cite{wang2023dynamic} incorporates the higher computational cost of higher resolution models in their optimization process by providing a fixed computation budget for each trial. Our proposed method optimizes model selection for task performance while accounting for computational cost by optimizing model complexity subject to a specifiable sub-optimality tolerance. Additionally, our method contributes spatially local resolution optimization, which allows some object regions to be higher resolution than others, while~\cite{wang2023dynamic} optimizes a global resolution parameter for a granular media manipulation task, which can also be made more computationally tractable using neural density fields rather than particle-based models~\cite{xue2023neural}. However, many common deformable objects such as clothes and rope, are better represented when some object regions have higher particle density than others~\cite{wang2023goal}. Our method informs tradeoffs between model complexity and task performance using the knowledge of semantically distinct regions of the object.

\par \textbf{Adaptive Mesh Simplification for Predictive Dynamics Models: }
Our strategies for adapting model resolution builds on existing methods that prioritize geometric and physical fidelity. The core objective of geometric simplification is to preserve spatial topology as much as possible while reducing mesh components~\cite{cignoni1998comparison, pfaff2020learning, fortunato2022multiscale}. Prediction error can also guide optimization of both spatial and temporal resolution such that complexity reductions minimally impact accuracy~\cite{liu2023fast, wu2023learning}. 
Meanwhile, other work in computational dynamics modeling use intelligent parallelizable multi-resolution message passing to reduce the number of sequential computation steps~\cite{cao2023efficient, janny2023eagle, perera2024multiscale}, which can be used in conjunction with mesh simplification~\cite{grigorev2023hood}. Since the priority in this paper is task performance, we can simplify more aggressively than those methods can. 

\par \textbf{Autonomous Tree Manipulation: }
We chose autonomous tree manipulation as a sample problem because recent advances in machine learning have enabled greater use of robots in agricultural settings, where branches can be intertwined and partially observable. To our knowledge, there is little existing work in model-based approaches, even though there can be benefits to managing occlusion and interwoven branches. Existing work to compute policies to move tree branches~\cite{kim2024towards, jacob2024gentle, jacob2024learning} simulate the entire tree, even for actions where only a small part is manipulated. The space of tree models is restricted for computational practicality, so we believe that task-specific mesh simplifications can bring us closer to practical model-based planning with trees.

\section{Problem Statement}


\newcommand{\sT}[2]{s_{#1, #2}}

\newcommand{\sRobotT}[1]{\sT{#1}{\robot}}
\newcommand{\sObjT}[1]{\sT{#1}{\obj}}
\newcommand{\obj}{\text{obj}}
\newcommand{\robot}{\text{robot}}

Our proposed method aims to automatically generate dynamics models for given planning queries that balance task performance with computational efficiency. 
We focus on planning queries where a robot end-effector manipulates an object to achieve a desired configuration. The robot uses MPC: computing a plan, which it then refines in response to new state information. While MPC can compensate for many modeling errors, some approximations significantly compromises the system's ability to reach goals. 

In this work, the computational complexity of the GNN dynamics model $z_{t+1} = \hat{f}(z_t, a_t)$ is determined by the number of components in a graph $z_t$, which is an encoding of state $s_t$. $s_t$ contains the 3D positions and velocities of both the end effector: $\sRobotT{t} = (x_t, \dot{x}_t)$ and $P$ points that represent the manipulated object: $\sObjT{t}$.
The action $a_t$ is an end-effector displacement. To inform region-specific model resolutions, we assume access to a segmentation of the point cloud in $s_t$. 
This frames the central challenge as \textit{automatically determining the minimal model complexity for successful task completion} given these regions. 

Task performance on a planning query $(s_1, s_G)$ is defined by the cost between final object state $s_{T}$ and desired goal object state $s_{G}$ after running MPC for $T$ time steps. 
The planning horizon is $T_{\text{horizon}}$ timesteps with replanning occurring every $T_{\text{replan}}$ steps. We assume full state observability. 
The cost function is the average distance of points that move more than $
\delta$ between $s_{1, \obj}$ and $s_{G, \obj}$.
Denoting point $i$ in $s_{t, \obj}$ is denoted as $s^i_{t, \obj}$, we define this mask:
\begin{equation}
m_i =  \mathbbm{1}{(|s^i_{1,\obj} - s^i_{G,\obj} | > \delta)}
\label{eq:mask}
\end{equation}

The cost function for the task is then:
\begin{equation}
\label{eq:model_gen_cost_function}
c(s, s_G) = \frac{1}{\sum_{i=1}m_i} \sum_{i=1}^P m_i||s^i_{t,\obj} -  s^i_{G,\obj}||
\end{equation}
Since the cost is only defined for states, planning with multiple graph resolutions requires our method to optimize a different function for planning. 

\newcommand{\buildgraph}{\textsc{BuildGraph}}
\newcommand{\omegafull}{\mathbf{1}^{dim(\modelres)}}
\newcommand{\fullres}{\mathbf{1}^{dim(\modelres)}}
\newcommand{\optimalres}{\mathbf{\vec{\omega}^*}}
\newcommand{\perftol}{\epsilon_{tol}}
\newcommand{\metamodel}{\hat{{\mathcal{F}}}(s,s_g)}

\section{Approach}
\textbf{Overview:} Our proposed method is to learn a model generator that given planning query $(s_1, s_G)$, outputs an efficient resolution for each region of the object, which we denote as $\modelres$. $\modelres$ parameterizes a simplified graph encoding of the state $z$, which is then used for planning. The model generator learns $\modelres$ on a dataset that reduces resolution while preserving task performance. 

Section~\ref{sec:model_space} defines a concrete space of dynamics models with region-specific resolutions. 
Section~\ref{sec:model_generator} then describes our model generator instantiation and how to train it from a dataset. 
Finally, Section~\ref{sec:model_res_data_collection} includes the optimization algorithm we use to construct the dataset of planning queries to efficient $\modelres$ on which the model generator is trained.

\subsection{Space of Dynamics Models}
A vector $\modelres$, which indicates the resolution for each region in $s_{t, 
obj}$ when building graph $z_t$, specifies the resolution of the GNN dynamics model. We denote constructing $z_t$ from $s_t$ as $\buildgraph(s_t, \modelres)$. 
For a point cloud with $K$ regions (i.e. segments), $\modelres$ is a $K$-dimensional vector where $\bomega_k$ indicates the resolution of region $k$ when building $z_t$. If $\bomega_k=1$, then a \emph{high}-resolution version of vertices in segment $k$ should be used. If $\bomega_k = 0$, then the region should be simplified to a \emph{low}-resolution with a single node.
$\hat{\Omega} := \{0,1\}^K$ then quantifies the space of dynamics models. 

\begin{figure*}
\includegraphics[width=\textwidth]{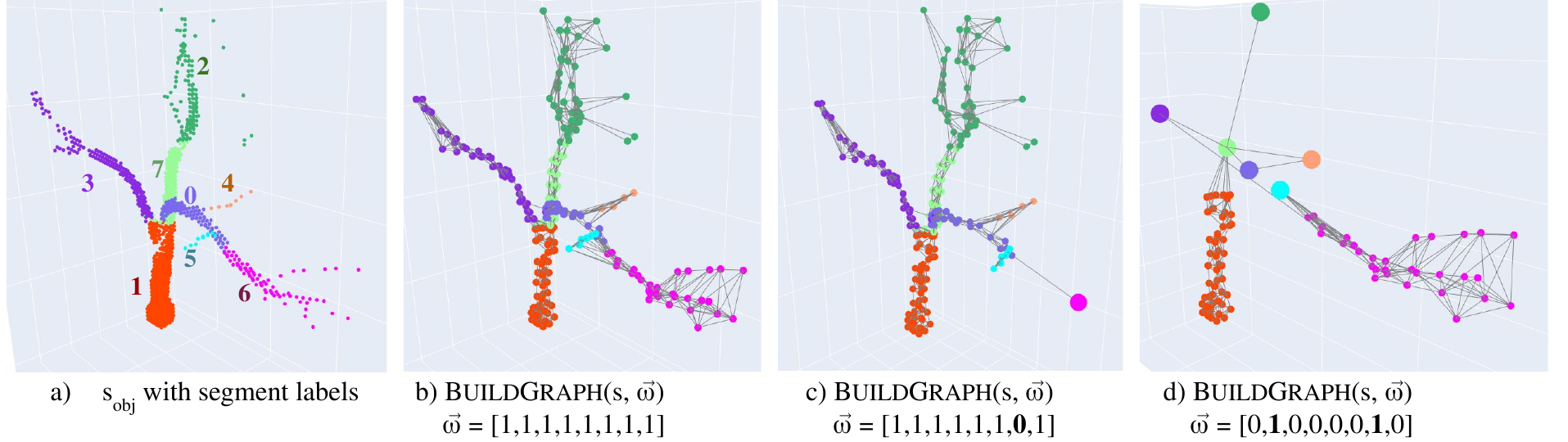}
        \caption{Examples of $z$ constructed from $\buildgraph$ from $\modelres$ and $s_{\obj}$. a) shows the input pointcloud with segments labeled, and b-d show meshes with segments simplified with different $\modelres$ }
        \label{fig:context_based_graph_simplification}
    \end{figure*}

Each graph $z_t$ is comprised of vertices $V_t$ and mesh edges $E_t$. Vertices represent a \emph{subset} of particles in $s_{t, 
obj}$ indexed by $i'$. Each vertex $v^{i'}_t \in V_t$ contains: 1) $i$, the index of its corresponding particle in $s_{t, \obj}$ 2) that particle's position position $v^{i', \mathrm{pos}}$, velocity history $v^{i', \mathrm{vel}}$, and segment $v^{i', \mathrm{seg}}$. Edge features contain distances between particles. The edge connections are computed by iteratively connecting k-NN until the graph is connected~\cite{qi20173d}. This method fits our domain, but there are general-purpose methods for mapping point clouds into a graph representation that we believe would work well~\cite{shi2020point}. 
Examples of outputs of $\buildgraph(s, \modelres)$ with different values of $\modelres$ and thus different regions simplified to a single vertex are shown in Fig.~\ref{fig:context_based_graph_simplification}.

The GNN dynamics model computes multi-step rollouts by chaining single step rollouts of the form $z_{t+1} \gets \hat{f}(z_t, a_t)$ where the initial $z_t$ comes from $\buildgraph(s_t, \modelres)$ and later $z_t'$ where $(t' > t)$ are computed from $z_{t'-1}$.
The complexity of the graph, and thus the computation time for computing scales quadratically with the number of graph components. The computational complexity of GNN evaluation scales quadratically with the number of vertices and edges in $z_t$.

The model is trained on a randomly generated dataset of 600K interactions in the environment with randomly selected $\modelres$. In this paper, we process all graphs with the $\textsc{Encode} \rightarrow \textsc{Process} \rightarrow \textsc{Decode}$ architecture. The node, edge, and global encoders each contain three fully-connected layers of size 128. The processor contains 10 layers, corresponding to 10 message passing steps. The decoder is usage-dependent; the dynamics model $\hat{f}(z_t, a_t)$ uses a 3-layer fully-connected decoder that outputs an acceleration for every vertex to compute the next state as a graph $z_{t+1}$ using the algorithm in~\cite{huang2022mesh, li2018learning}. In this paradigm, $\hat{f}$ can operate on graphs of very different sizes. Actions, which are 3D displacement vectors, are appended as vertex features.

\label{sec:model_space}
\subsection{Query-Guided Model Generator}
\label{sec:model_generator}
\begin{figure*}[h]
\includegraphics[width=\textwidth]{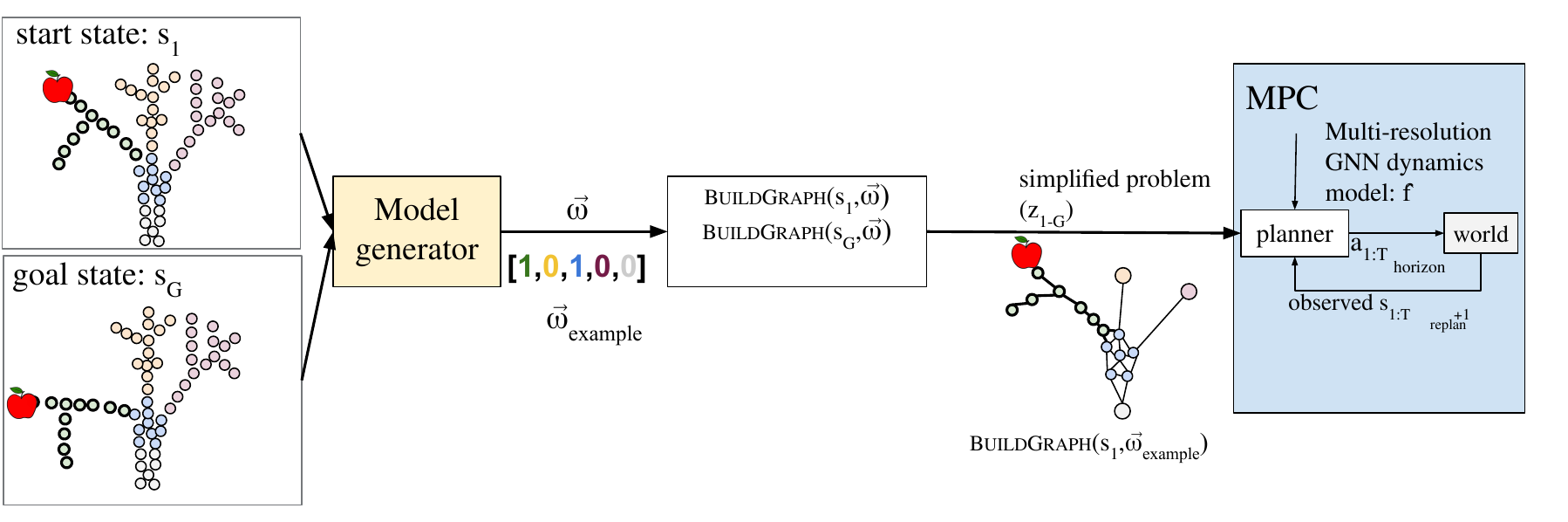}
\caption{Overview of how a model generator is used. Starting at the left, start and goal states with the object represented as pointclouds are inputted to the model generator, which predicts a resolution vector $\modelres$, indicating which regions to model in high vs. low resolution. This informs an encoder to produce a simplified planning query, $(z_{1-G})$, which is solved using model predictive control (MPC). The first $T_{\text{replan}}$ actions are executed before replanning. }
\label{fig:model_gen_visual_overview}
\end{figure*}

This next section describes the model generator: its definition, implementation, and how we train it. 
As illustrated in Fig.~\ref{fig:model_gen_visual_overview}, the planning query is specified by encoding the start state $s_1$ and goal state $s_G$ into a single graph $z_{1-G} = (V_{1-G}, E_{1-G})$. First, the start and goal state are encoded separately using $\modelres = \omegafull$: $z_1 = \buildgraph(s_1, \omegafull)$ and $z_G = \buildgraph(s_G, \omegafull)$. $V_{1-G}$ is the union of vertices in $z_1$ and $z_G$. The edges connect vertices from $z_1$ to the corresponding vertex in $z_G$ with the displacement and direction as attributes. 

The model generator learns a distribution of model resolutions $p_{\theta}(\modelres | z_{1-G})$, which it can then sample from at test time. The dataset $\mathcal{D}$ is comprised of $s_t, s_G, \optimalres$ tuples. Since the distribution of $\optimalres$ can be multimodal, we use a diffusion model to approximate a distribution over $\modelres$: $p_{\theta}(\modelres | z_{1-G})$. We use the noise scheduler and loss function approximation from Denoising Diffusion Probabilistic Models~\cite{ho2020denoising} with a slightly modified parameterization $\epsilon_{\theta}(\modelres, s_{1}, s_{G}, t)$, described in Algorithm~\ref{alg:epsilontheta} which uses the encoder and processor GNN to denoise $z_{1-G}$. The global vector contains only the diffusion process timestep $t$. 

\begin{algorithm}
	\caption{Procedure for $\epsilon_{\theta}(\modelres, s_1, s_G, t)$}

\begin{algorithmic}[1]
	\State \textbf{Input:} $(\modelres, s_{1}, s_{G}, t)$
	\State \textbf{Output:} $\modelres'$ 
	\State $z_1 = \buildgraph(s_1, \fullres)$ \Comment{Using full resolution}
	\State $z_G = \buildgraph(s_G, \fullres)$ 
	\State $V_{1-G} = (V_1, V_G)$
	\State Construct $E_{1-G}$ by connecting corresponding vertices in $V_{1-G}$ with their relative positions and direction as edge attributes
	\State $z_{1-G} = (V_{1-G}, E_{1-G})$ 
	\State Initialize empty $V_{input}$
	\For{each vertex \( v_i \in V_{1-G} \)}
		\State $v_{input,i} = v^{pos}_i | \modelres_{v^{seg}_i}$ \Comment{Concatenate position and resolution for segment}
        	\State Add $v_{input,i}$ to $V_{input}$
	\EndFor
	\State $V'_{input}, E'_{1-G} = GNN(V_{input}, E_{1-G})$ \Comment{GNN encoder and processor where final node embedding is a scalar for each node }
	\For{segment $k=1, \dots, K$}
	    \State $V'_{k} = \{v_i \ V'_{input} \; | \; v_i^{seg} = k\} $
	    \State $\bomega'_j =  \frac{1}{|V'_k|} \sum_{v_i \in V'_k} v_i$ 
	\EndFor
\end{algorithmic}
\label{alg:epsilontheta}
\end{algorithm}

\newcommand{\alphat}{\bar{\alpha_t}}

\textbf{Training:} During each training step on datapoint $s_{1}, s_{G}, \modelres*$, we first sample $t \sim Uniform([1,\ldots T])$, $\epsilon \sim \mathcal{N}(0,I)$ and $\modelres_0 \sim q(2\modelres*-1)$ using the noise scheduler in~\cite{ho2020denoising}, normalizing $\optimalres$. Then, we optimize the parameters of $\epsilon_{\theta}$ with the simplified loss $\nabla_{\theta}||\epsilon-\epsilon_{\theta}(\sqrt{\alphat}\modelres_0 + \sqrt{1-\alphat}\epsilon,s_{1}, s_{G}, t)||^2$ until the validation loss begins to increase. $\alpha_t$ and $\alphat$ are computed with the DDPM noise scheduler. We train on a dataset of 6075 samples holding out 10\% for validation. 
\\
\textbf{Sampling:} At test time, the model generator samples $\modelres$ for each new $s_{1}, s_G$ by performing $T=100$ denoising steps from $T$ to $1$. At each step, random noise $\mathbf{x} \sim \mathcal{N}(0,1)$ is denoised using the update step shown in Equation~\ref{eq:denoising_step}. 
The final outputted $\modelres$ is denormalized.

\begin{equation}
\modelres_{t-1} = \frac{1}{\sqrt{\alpha_t}} \left(\modelres_{t} - \frac{1-\alpha_t}{\sqrt{1-\alpha_t}}\epsilon_{\theta}(\modelres, s_1, s_G, t) \right) + \sigma_t\mathbf{x} 
\label{eq:denoising_step}
\end{equation}

We denote the end-to-end sampling process as $\modelres = \mathcal{M}_{\theta}(s_1, s_G)$.

The model generator is trained on a dataset of 11K $s_1, s_G, \optimalres$ tuples. We optimize using DDPM with a cosine $\beta$ schedule in batches of 64 until the validation loss begins to increase. Section~\ref{sec:model_res_data_collection} describes how to collect this dataset.

\subsection{Using a Model Generator in Planning}
\textbf{Usage during planning:} During MPC, the predicted $\optimalres$ is sampled: $\modelres \sim \mathcal{M}(s_1, s_T)$ to construct $z_1$ with efficient resolution, which is then used by the dynamics model for rollouts. $\modelres$ can be kept the same or resampled during later planning iterations by using $s_t$ as the start state input to $\mathcal{M}_{\theta}$.

Since the dynamics model inputs and outputs multi-resolution graphs rather than states, we use a form of a chamfer distance~\cite{fan2017point} cost function \emph{for planning} to account for varying numbers of vertices in $z_t$. 
\newcommand{\trajdist}{c_{plan}} $\trajdist(z_t, z_{G}) = \sum_{i'=1}^P m_{i'} \min_{j' < P \, | \, m_{j'} } \|x^{i'}_{1,\obj} - x^{j'}_{G,\obj}\|^2 + \sum_{j'=1}^P m_{j'} \min_{i<P\, |\, m_{i'}} \|x^{i'}_{1,\obj} - x^{j'}_{G,\obj}\|^2$.

\subsection{Dataset Construction via Chained Optimization}
\label{sec:model_res_data_collection}
\newcommand{\dyntol}{w^{dyn}}
This section outlines our approach for generating a resolution $\optimalres$ which in order to construct a dataset mapping planning queries as $(s_1, s_G)$ pairs to an optimized $\modelres$. 

This section describes our method for computing a dataset $\mathcal{D}$ of particular planning queries $(s_1, s_G)$ and an optimal resolution $\optimalres$ that balances computation time and task performance. 

The challenge is that the objective we care about the most is \textit{closed-loop task performance}, yet interleaved planning and execution is expensive. For each candidate $\modelres$, we must run complete planning episodes including planning, execution, observing state changes, and re-planning. 

\begin{figure}
\includegraphics[width=0.5\textwidth]{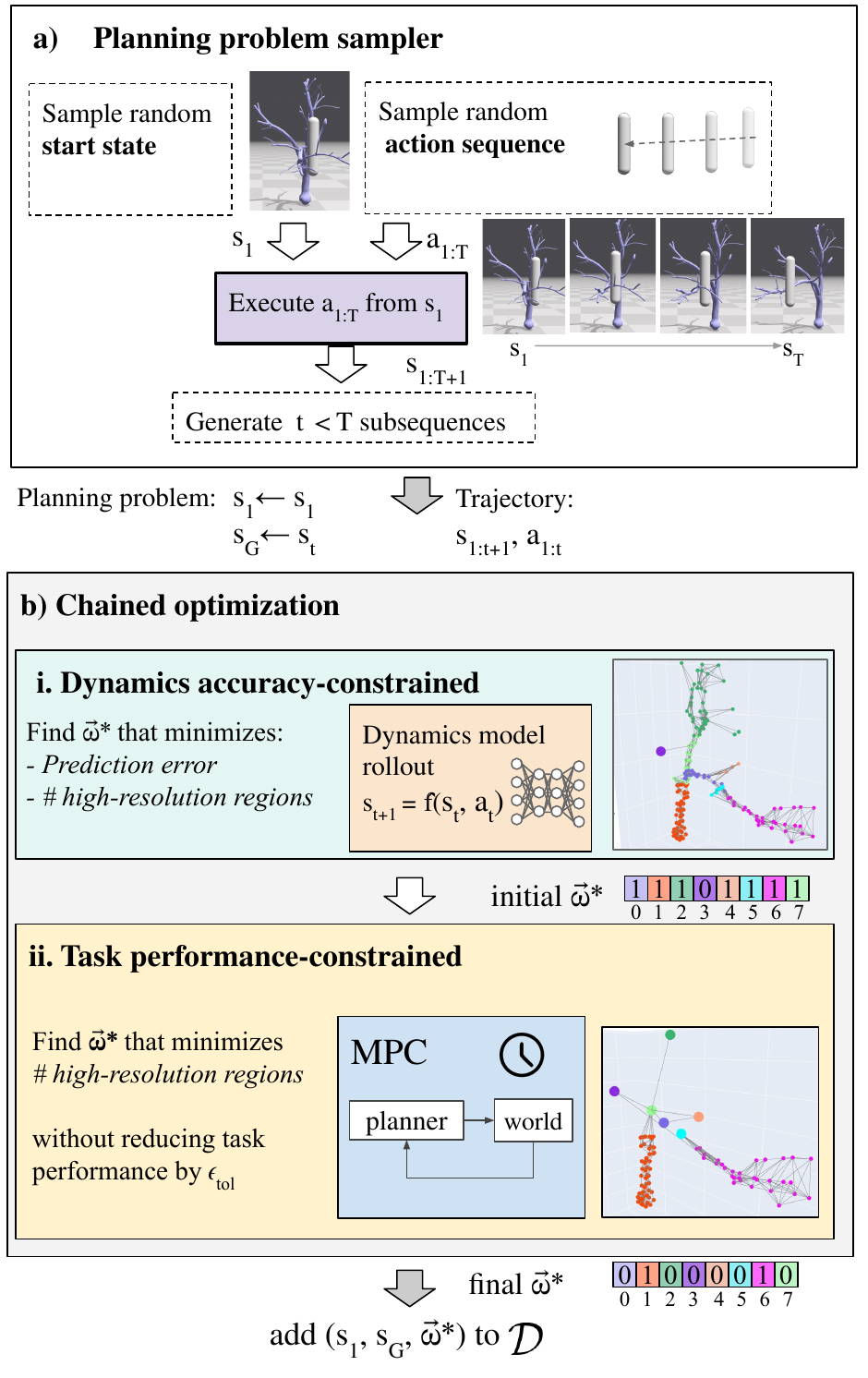}
\caption{
Data flow diagram of chained optimization algorithm used to generate $\optimalres$ for a planning query $(s_1, s_G)$. A random action sequence is executed from $s_1$ from which planning queries are formed from trajectory rollouts. The chained optimization has two stages shown with a visual example. Region numbers and colors correspond to regions (see Fig.~\ref{fig:context_based_graph_simplification} for key). The first optimization stage uses dynamics model accuracy to initialize $\optimalres$ for the second stage, which further simplifies it through a task-performance objective. Each $(s_1, s_G) \rightarrow \optimalres$ is added to $\mathcal{D}$.}
\label{fig:chained_optimization}
\end{figure}

Our key strategy for making the dataset construction tractable is to employ a two-stage optimization strategy. Fig.~\ref{fig:chained_optimization} shows the data flow. The first stage (Sec.~\ref{sec:dynamics_optimization}) reduces $\modelres$ to a good initialization using a dynamics accuracy loss, which can be quickly evaluated through highly parallelizable forward rollouts. The second stage (Sec.~\ref{sec:mpc_optimization}) further reduces $\modelres$ using closed-loop task performance, which is expensive to evaluate but directly measures our true objective.

$(s_1, s_G)$ are sampled from a random planning query distribution consisting of reachable goal configurations and start states. $\sRobotT{0}$ is sampled from random collision-free positions. Reachable $s_G$ are achieved by moving the end-effector in random motions and saving ($s_1$, $s_G$) where the tree moves without becoming unstable.

\subsubsection{Dynamics Model Accuracy-Constrained Model Simplification}
\label{sec:dynamics_optimization}

Dynamics \textit{accuracy}-constrained model simplification initializes $\optimalres$ by minimizing the number of high-resolution regions ($\sum \modelres$) subject to a constraint in dynamics accuracy, $\dyntol$. We do this by solving the following optimization problem:

\begin{equation}
\begin{aligned}
    \optimalres = & \underset{\modelres \in \hat{\Omega}}{\text{min}} \; 
 \trajdist \left( \textsc{BuildGraph}(s_T,\modelres),z_G \right)+\dyntol |\modelres|_1 \\
     \text{subject to} \quad  &\mathrm{max}(\modelres) \leq 1 \\
     &\mathrm{min} (\modelres) \geq 0 \\
\end{aligned}
\label{eq:dyn_opt}
\end{equation}

Here, $\textsc{BuildGraph}(s_T, \modelres)$ computes a rollout using a model simplified according to $\modelres$, and $\trajdist$ measures the distance between the predicted and observed trajectories. We use CMA-ES to solve this optimization problem, with a conservative dynamics tolerance $\dyntol = 0.005$, mean initialized to 0.7, and population size $\lambda = 20$ for our experiments. 

\subsubsection{Closed-Loop Task Performance-Constrained Optimization}
\label{sec:mpc_optimization}
In this second stage, we further reduce $\optimalres$ by evaluating closed-loop performance. We seek the most simplified model (i.e., most simplified regions) that when used for MPC, achieve a cost within $\perftol$ of the best observed performance for that planning query:

\begin{equation}
\begin{aligned}
    & \optimalres \gets \underset{\modelres \in \hat{\Omega}}{\text{min}} \quad \sum_{k=1}^N \modelres_k \\
    & \text{subject to} \quad c^{\modelres}(s_T, s_G) \leq c^* + \perftol \\
\end{aligned}
\label{eq:plan_opt_obj}
\end{equation}

Here, $c^{\modelres}(s_T, s_G)$ is the result of planning using $\modelres$ as input to the encoder and $c^*$ is the best cost observed so far during optimization. Importantly, costs are measured in the true environment state, not the simplified encoded space. We do not assume dynamics model accuracy, and the planner reasons over the encoded (simplified) state, which differs from the ground-truth.

The optimization process (Algorithm.~\ref{alg:mpc_optimization}) begins with an initial $\modelres_0$ from the previous stage. At each iteration $i$, we evaluate $\hat{\omega}_{i}$ via MPC, and attempt to further simplify the model for $\modelres_{i+1}$. The process continues until no further reduction is possible without violating the performance threshold.

\begin{algorithm}
\caption{}
\begin{algorithmic}[1]
\State \textbf{Input:} Planning query $(s_1, s_G)$, prior $\optimalres$, $I_{\text{grace}}$ grace period, $K_{\text{max}}$ max number regions to simplify each round.
\State \textbf{Output:} $\optimalres$
\State $i \gets 0$
\State $\modelres^0 \gets \optimalres$ \Comment {Initialize using prior}
\While{$\optimalres$ unchanged $\geq I_{\text{grace}}$ iterations or $\sum \optimalres_t = 0$}
    \Comment{Run MPC using $\modelres^i$}
    \State $c^i \gets \texttt{MPC}(s_1, s_G, \modelres^i)$
    \If{$i=1$ or $c_i <  c^*$} 
        \State $c^* \gets c^i$ \Comment{Update optimal cost seen so far}
    \EndIf
    \If{$c^i \leq c^* + \perftol$}
        \State $\optimalres \gets \modelres_i$
        \State increase $K_{\text{max}}$ 
    \Else
        \State decrease $K_{\text{max}}$ 
    \EndIf
    \State $\modelres^{i+1} \gets \modelres^{i}$ with $K \leq K_{\text{max}}$ random indices set to 0.
    \State $i \gets i + 1$
\EndWhile
\State \Return $\hat{\omega}^*$
\end{algorithmic}
\label{alg:mpc_optimization}
\end{algorithm}

\section{Experiments}


We evaluate our task-guided model generation approach on a tree manipulation task, analyzing both learned model resolutions and end-to-end task performance. Tree manipulation is a good testbed due to both computationally intensive state representations and complex branch interdependencies.

Section~\ref{sec:model_gen_experimental_setup} section describes the experimental setup including task description and planner configuration. Section~\ref{sec:model_gen_distribution} then analyzes the outputs of the model generator, specifically the distribution of selected resolutions for particular planning queries. Section~\ref{sec:model_gen_quantitative} then compares the cost and planning time using our model generator to baseline dynamics models. %

\newcommand{\minisec}[1]{\par\vspace{0.5\baselineskip}\noindent\textbf{#1:}\space}
\subsection{Experimental Setup}
\label{sec:model_gen_experimental_setup}
\newcommand{\treepanelwidth}{0.23\textwidth}
\begin{figure}
    \centering
    \includegraphics[width=\treepanelwidth]{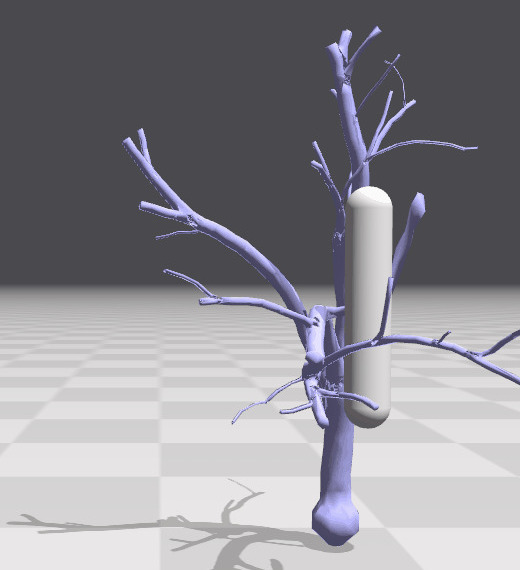}%
    \hfill
    \includegraphics[width=\treepanelwidth]{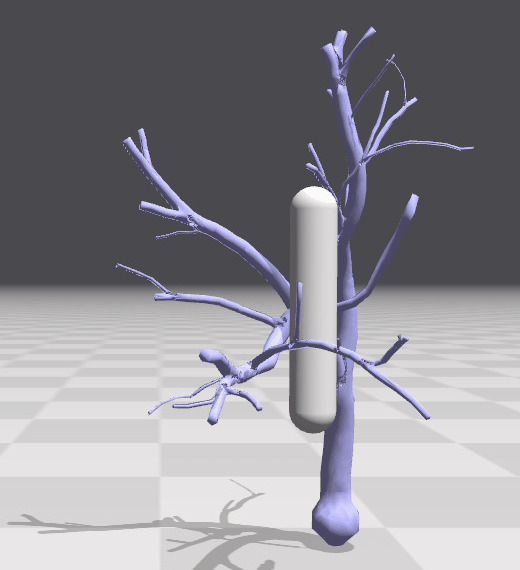}%
    \hfill
    \includegraphics[width=\treepanelwidth]{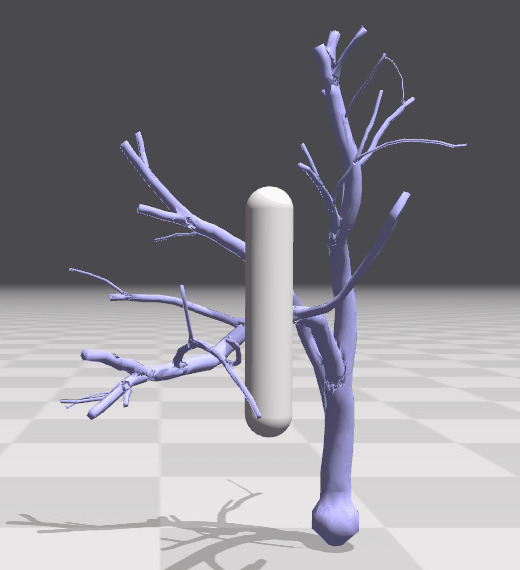}%
    \hfill
    \includegraphics[width=\treepanelwidth]{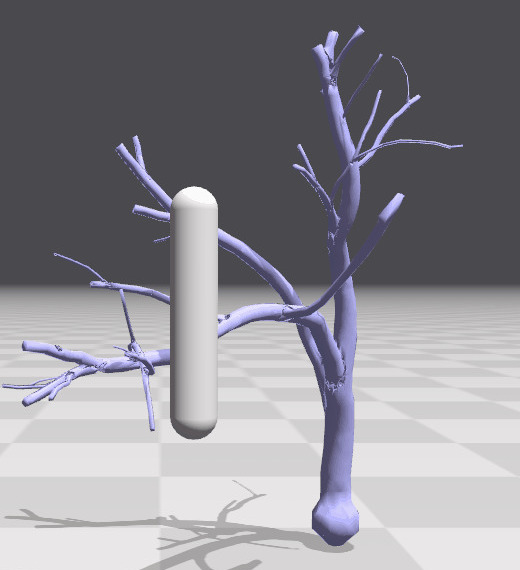}%
    \hfill
    \caption{Frames showing an example of the free-floating capsule end effector pushing a branch in our SoftGym environment.}
    \label{fig:treeenv}
\end{figure}

\minisec{Task and environment}
The task is to move a branch to a target configuration using a large cylindrical non-prehensile end-effector~\cite{lee2024sonicboom}.
Our simulated version of the end-effector and tree is implemented in a custom environment built in SoftGym~\cite{lin2021softgym} shown in Fig.~\ref{fig:treeenv}. Although the robot is represented with a free floating end-effector, in practice, we can use inverse kinematics to generate joint-space trajectories that follow the end-effector trajectories. The tree is represented by 1317 particles from an apple tree generated by Grove3D~\cite{thegrove3d}. We implement realistic tree physics by customizing the Flex SoftBody simulator. 

$s_{\text{obs}}$ consists of all particles from the mesh. The end-effector state $(x_t,y_t,z_t)$, including the last 5 velocity commands, $(\dot{x}_{t:t-5}, \dot{y}_{t:t-5}, \dot{z}_{t:t-5})$, is also part of $s_{t}$. 
The end-effector is represented as a SoftGym capsule that can interact with the tree by pushing, but cannot attach or ``pick'' particles the same way end-effectors in other SoftGym tasks can. 

\minisec{Planning query generation}
We generate feasible planning queries by sampling random 80-step trajectories that move the end-effector toward random tree particles in random directions. Multiple planning queries come from each trajectory using different cut-off indices: $T_[\text{window}]$. The goal is $s_{T_{\text{window}}}$ with $24 \leq T_{\text{window}} \leq 72$. 5500 $(s_1, s_G, \optimalres)$ tuples were in the dataset.

\minisec{Planning algorithm}
We use MPPI as the planner~\cite{williams2016aggressive}. MPPI finds an optimal trajectory with a horizon $T_{horizon}$ = 32.
The trajectory cost is a weighted version of $c_{plan}(z_t, z_G)$:$\sum_{t'=1}^{H} \gamma^{t-H} \trajdist(z_{t'}, z_G)$ where $\gamma=0.95$.

MPPI iteratively computes a sequence of actions that minimize the predicted trajectory cost every $T_{\text{replan}}$ steps. 
At each time step $t$, the dynamics model predicts trajectories for a batch of randomly initialized action sequences: $\hat{s}_{t:H} \gets \hat{f}(s_t, a_{t:t+H-1}))$. Standard MPPI updates actions using the predicted trajectory costs. Hyperparameters are shown in Table~\ref{tab:mppi_params}

\begin{table}[h]
\centering
\caption{MPPI parameters}
\begin{tabular}{lr}
parameter & value \\
\hline
$T_{horizon}$ & 32 \\
\# samples & 64 \\
$T_{replan}$ & 4 \\
$\gamma$ & 0.95 \\
refinement steps (initial/other) & 5/3 \\
noise scale & $3.6 * 10^{-5}$
\end{tabular}
\label{tab:mppi_params}
\end{table}

\subsection{Optimal Model Resolution Distribution Analysis}
\label{sec:model_gen_distribution}

Here, we analyze whether our method generates query-guided model resolutions by comparing the distribution of the optimized resolutions and the distribution of learned resolutions on sets of similar planning queries. 

The top of Fig.~\ref{fig:distributions} shows the distribution of $p_{\theta}(\modelres)$ for planning queries categorized by which regions move the most between the start and goal configuration. The results indicate that $\optimalres$ differs substantially across tasks: different tasks require different regions to be represented in high resolution. The bottom of Fig.~\ref{fig:distributions} shows that the predictions from our diffusion model match the expected distribution for different classes of planning queries. 

Although we hypothesized that regions that move significantly between the start and the goal would more frequently appear in high resolution in $\optimalres$, the data suggests the trend generally holds, but the relationship is more nuanced. 
For example, the top of Fig.~\ref{fig:distributions}a shows which regions are included in $\optimalres$ for planning queries involving the branch highlighted in yellow, which corresponds to regions 0, 4, 5, and 6 shown in Fig.~\ref{fig:context_based_graph_simplification}. All segments are represented in high resolution at least once in $\optimalres$ for these queries, though region 6 (at the end of the branch) is by far the most common. Regions 0, 4, and 5 are commonly simplified despite being moved. Our diffusion model shows a similar distribution.

In contrast, for planning queries where only region 3 is moved (Fig.~\ref{fig:distributions}b), region 6 is rarely represented in high resolution in $\optimalres$. As expected, region 3 often in high resolution in $\optimalres$. However, region 2, the top of the tree, is often high resolution in $\optimalres$ as well. This coupling likely comes from message passing between interconnected segments. Fig.~\ref{fig:distributions}c shows similar coupling between regions 2 and 3.

\newcommand{\trajTypeWidth}{0.16\textwidth}
\newcommand{\trajTypeImgWidth}{\textwidth}
\begin{figure}
    \centering
       \begin{subfigure}[b]{\trajTypeWidth}
       \centering
      \includegraphics[width=\trajTypeImgWidth]{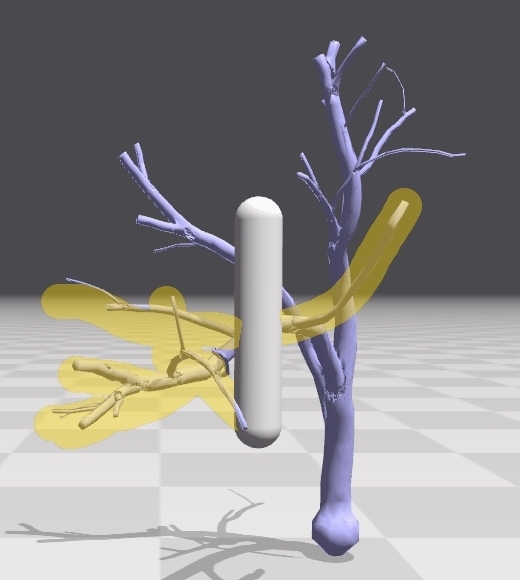} 
      \label{subfig:antlerpic}
     \end{subfigure}%
     \hfill
    \begin{subfigure}[b]{\trajTypeWidth}
       \centering
      \includegraphics[width=\trajTypeImgWidth]{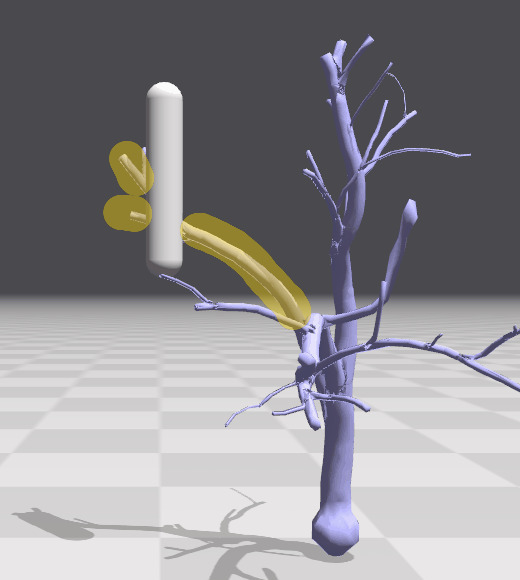} 
      \label{subfig:longpic}
     \end{subfigure}%
     \hfill
    \begin{subfigure}[b]{\trajTypeWidth}
       \centering
      \includegraphics[width=\trajTypeImgWidth]{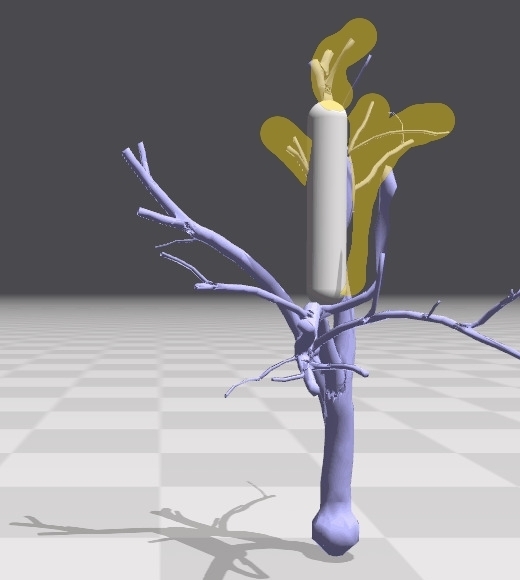} 
      \label{subfig:toppic}
     \end{subfigure}
       \begin{subfigure}[b]{\trajTypeWidth}
      \includegraphics[width=\textwidth]{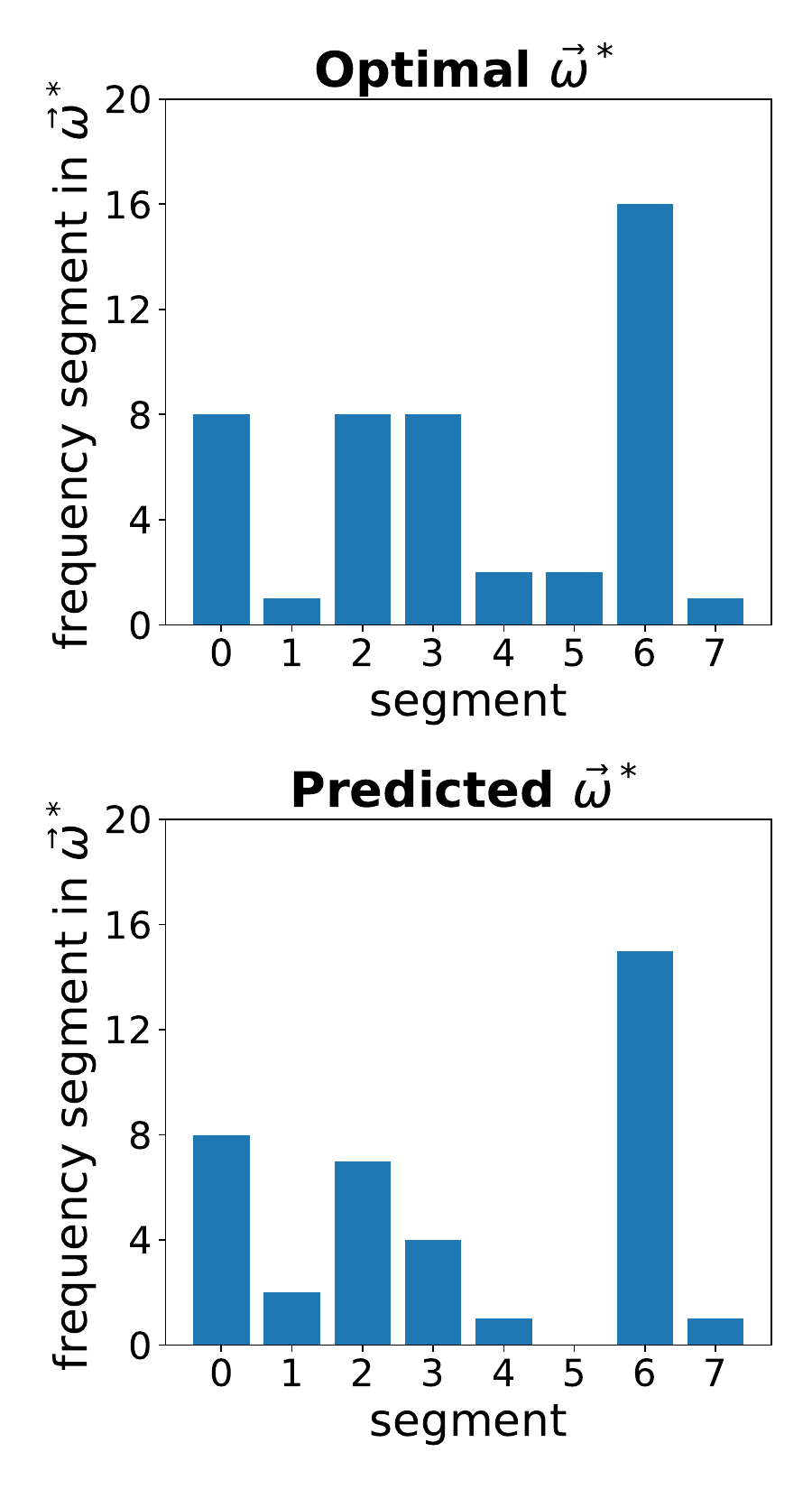} 
      \caption{}
      \label{subfig:antler_dist}
     \end{subfigure}%
     \hfill
       \begin{subfigure}[b]{\trajTypeWidth}
      \includegraphics[width=\linewidth]{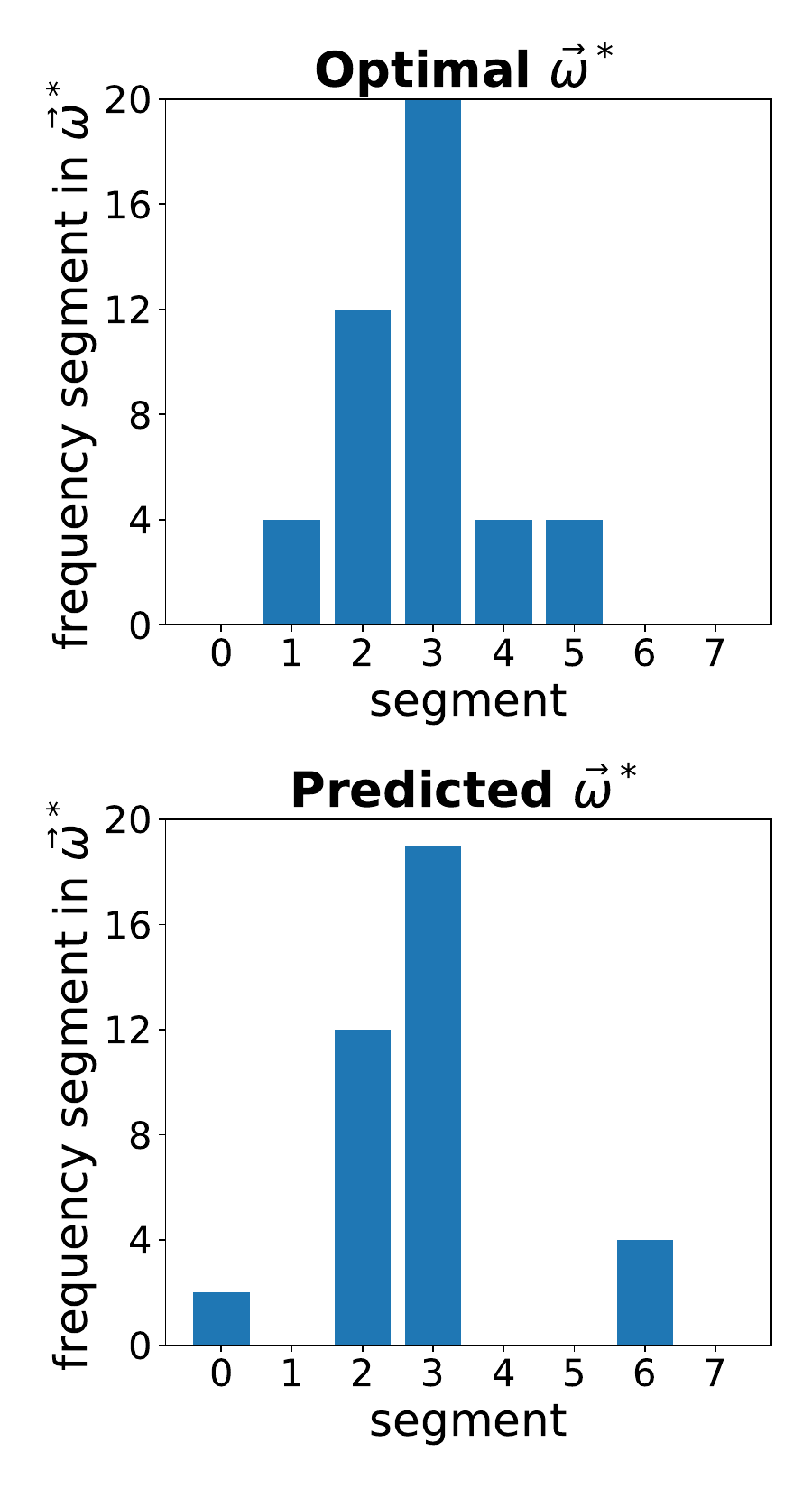} 
      \caption{}
      \label{subfig:long_dist}
     \end{subfigure}%
     \hfill
    \begin{subfigure}[b]{\trajTypeWidth}
      \includegraphics[width=\linewidth]{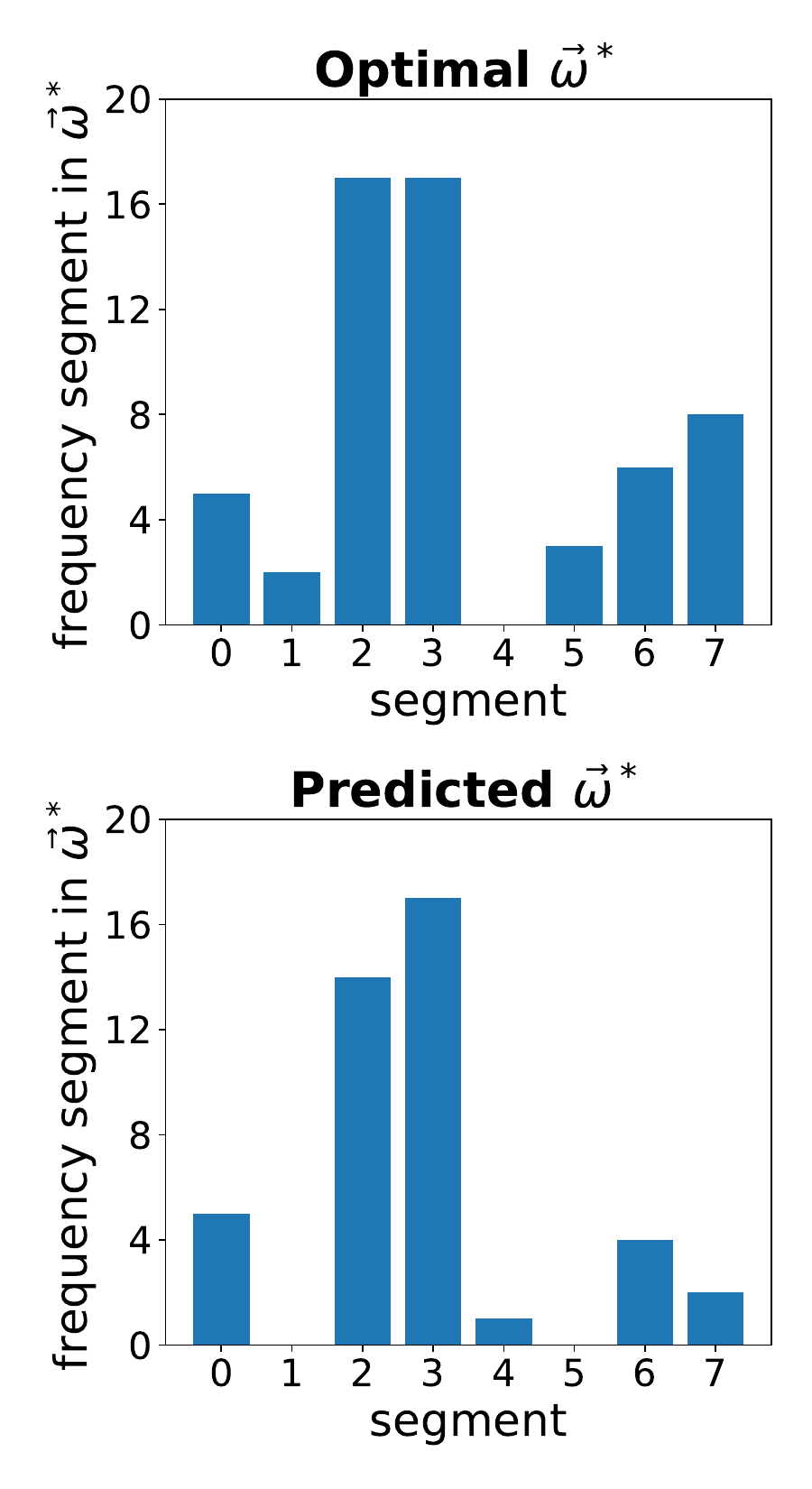} 
      \caption{}
      \label{subfig:topdist}
     \end{subfigure}%
    \caption{Distribution of selected model resolutions for three classes of planning queries. (Top) Branches that move in each planning query class are highlighted in yellow.  (Bottom) Frequency that each segment is represented in high resolution in the optimal resolution. Our diffusion model closely matches the target distribution.}
    \label{fig:distributions}
\end{figure}

\subsection{Task Performance Results}
\label{sec:model_gen_quantitative}
This section evaluates how task-specific model resolutions affect both task performance and planning time. We evaluate planning using our learned diffusion model compared to baselines on a set of 100 test queries not seen during training. 

\minisec{Baselines}
We compare against three baselines: \textbf{\texttt{full}} ($\modelres = \mathbf{1}^{dim(\modelres)}$), \textbf{\texttt{minimal}} ($\modelres = \mathbf{0}^{dim(\modelres)}$) and \texttt{\textbf{mode}} (most frequent $\optimalres$ in the training data, regardless of planning query)

\texttt{full} provides an upper bound on time, since it is equivalent to planning with the full resolution. \texttt{minimal} simplifies each segment, which is the fastest model, but also the coarsest. \texttt{mode} tests the extent to which query-guided selection helps performance by comparing it to always selecting the most common $\optimalres$ in $\mathcal{D}$.

\minisec{Evaluation metrics}
We measure task performance and planning time. 
Planning time includes the entire MPPI optimization process, including encoding and overhead from using the model generator (where applicable). Task performance is measured by the task objective from Eq.~\ref{eq:model_gen_cost_function}. All experiments are run on a desktop workstation with an AMD EPYC 7542P 32-Core Processor and NVIDIA RTX A6000 GPU. 

\begin{figure}
\begin{subfigure}[b]{\figwidth}
        \centering
        \includegraphics[width=\textwidth]{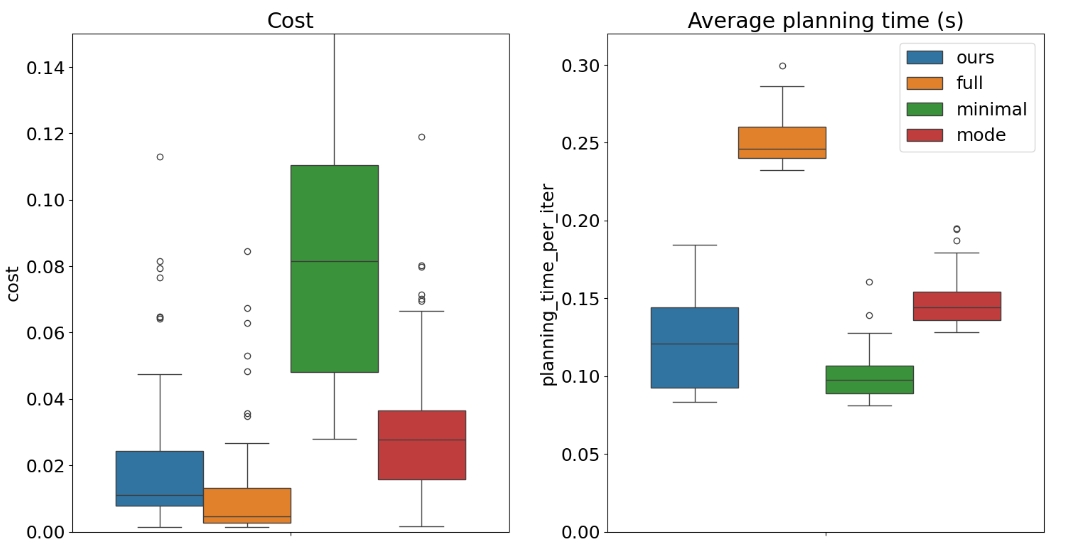}
        \caption{Box plot showing the costs (left) and average planning times (right) for our method compared to baselines. Outliers are shown with circles.}
        \label{fig:model_gen_e2e}
\end{subfigure}
%
        \caption{Mean (standard deviation) for final cost and average planning time for our algorithm and baselines. }
        \label{fig:baselines}
\end{figure}
Figure~\ref{fig:model_gen_e2e} shows the end-to-end performance of planning with our model generator compared to the baseline dynamics models. 
Our main finding is that compared to \texttt{full}, our query-guided model generator achieves a 2.1x speedup in planning time at a slightly lower but comparable task performance. The observed 0.006 gap in cost corresponds with about a 1 cm average increase in average node distance for particles that significantly moved. While \texttt{minimal} is even faster, it produces unacceptably high costs due to excessive model simplification. Selecting $\optimalres$ using the \texttt{mode} baseline (independent of $(s_1, s_G)$) results in both higher costs and slower planning than from our model generator. 




Overall, the results demonstrate the potential for learned model generators to balance planning performance and computational efficiency, even with highly multi-modal optimal model distributions. The computational benefits are modest in our current setup, since we downsample from 1317 points in $s_{\text{obs}}$ to 227 for regions at full resolution ($\modelres = \mathbf{1}^8$). Greater speedups would likely emerge from applying this approach to larger pointclouds.

\section{Limitations and Future Work}
The capabilities of our proposed model generator can be significantly higher by using pre-existing knowledge, especially representations and object relationships. 
For example, general purpose deformable object descriptions that leverage semantic information~\cite{wang2023dynamic} help transfer knowledge about world models and task specifications between tasks. 
Future work can facilitate this transfer by learning the regions used for simplification from a general-purpose input such as a point cloud. 

Another limitation is the noisy relationship between closed-loop performance and resolution, a problem also faced by~\cite{wang2023dynamic}. High-quality differentiable simulators for deformable objects~\cite{chen2024differentiable}, which also enable end-to-end policy optimization, can provide a more direct gradient signal. 

\section{Conclusion}
This paper presents an approach for generating task-informed dynamics models that simplify geometric regions of deformable objects to balance task performance and computational efficiency.
To compute the dataset of optimal region-specific resolutions, we propose a two-stage optimization method. Our experiments show that the model generator produces resolutions that significantly reduce planning time while maintaining task performance compared to several task-agnostic resolution baselines. We believe our model generator and local resolution strategy can serve as a foundation for transforming complex state data into more efficient task-oriented representations. 

\bibliographystyle{ieeetr}
\bibliography{references/active_mde_references, references/references, references/planorparam, references/model_gen_reference, references/modelPreconditions}

\end{document}